\documentclass[conference]{IEEEtran}
\usepackage{times}

\usepackage[numbers,sort&compress]{natbib}

\usepackage{multicol}
\usepackage[bookmarks=true]{hyperref}
\usepackage{graphicx}

\usepackage{amsmath,amsfonts}
\usepackage{algorithmic}
\usepackage{array}
\usepackage[caption=false,font=normalsize,labelfont=sf,textfont=sf]{subfig}
\usepackage{textcomp}
\usepackage{stfloats}
\usepackage{url}
\usepackage{verbatim}
\usepackage{amssymb}

\begin{document}

\title{Relaxation-Aware Multimodal Sensing of Soft Gripper Driven by Structure-Perception-Learning}

\author{
\authorblockN{
Yanzhe Wang\textsuperscript{1,2,\authorrefmark{1}},
Hao Wu\textsuperscript{1,2,\authorrefmark{1}},
Ziyi Zheng\textsuperscript{1,2},
and Huixu Dong\textsuperscript{1,2,\authorrefmark{2}}
}

\authorblockA{
\textsuperscript{1}Grasp Lab, School of Mechanical Engineering, Zhejiang University
\quad
\textsuperscript{2}Torch Kernel Co., Ltd.
}
\authorblockA{
\textsuperscript{\authorrefmark{1}}These authors contributed equally to this work.
\quad
\textsuperscript{\authorrefmark{2}}Corresponding author: huixudong@zju.edu.cn
}
}

\maketitle

\begin{abstract}
Achieving stable, sustained grasping with soft robotic hands remains a fundamental challenge. Compliance enables safe and adaptive contact, yet the intrinsic viscoelasticity of soft polymers leads to stress relaxation and a continuous decay of grasping force during holding. Inspired by human grasping, which combines phase-dependent stiffness regulation with continuous sensing and feedback, this paper presents an integrated structure--perception--learning framework. We develop a variable-stiffness soft gripper that uses onboard vision and infrared thermography to track deformation and the temperature field in real time, preserving continuous tracking of the interaction state. To mitigate relaxation-induced force decay, we propose a temperature-coupled viscoelastic force representation, together with a physics-informed learning model, to reconstruct the force trend and provide explicit compensation during holding. Experiments show that, in a 280s force-controlled grasp-and-hold task, the proposed method maintains the desired force with a mean absolute error of 0.066N, outperforming fixed-aperture and instantaneous-only baselines by 80\% and 95\%, respectively. Overall, the results support a mechanism--AI co-design view: mechanisms shape feasible interactions, while learning compensates remaining uncertainty in viscoelastic dynamics, together enabling stable, sustained grasping.

\end{abstract}

\IEEEpeerreviewmaketitle

\section{Introduction}

Achieving stable robotic manipulation in unstructured environments is fundamentally not limited by contact acquisition, but by the ability to maintain reliable and controllable interaction forces over extended time horizons.
Human manual manipulation provides a mature and efficient paradigm in this regard: stable grasping does not rely on a single sensory feedback channel, but instead emerges from the coordination of multiple capabilities.
Adjustable musculature enables phase-dependent stiffness regulation. Multimodal perception continuously monitors the evolving contact state. Higher-level regulatory mechanisms compensate force variations induced by deformation, fatigue, or external disturbances during prolonged holding. Together, these capabilities give the human hand a key property: continuous observability of interaction states as they evolve over time.

\begin{figure}[t]
    \centering
    \includegraphics[width=1\linewidth]{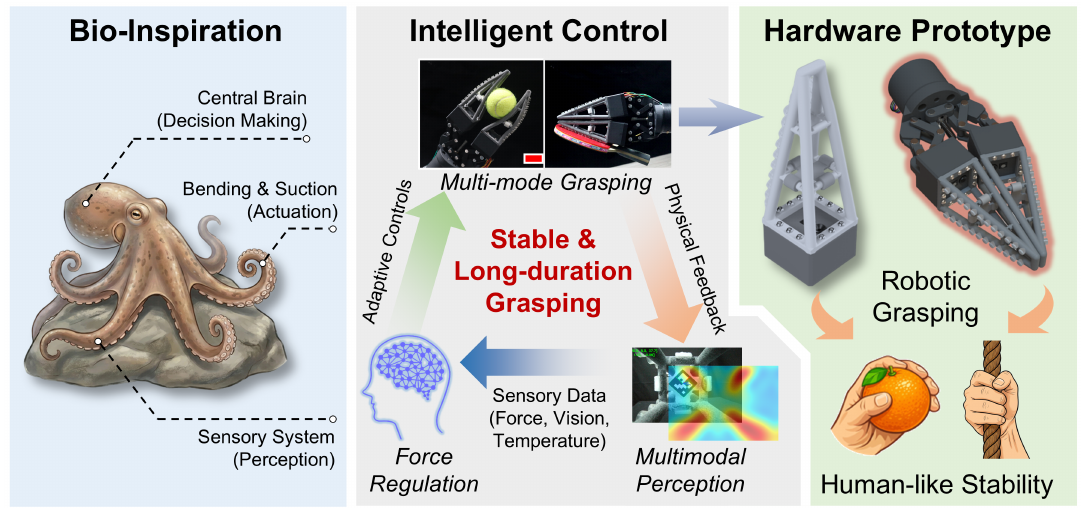}
\caption{
Inspired by the compliant bending and adhesive interaction mechanisms of octopus tentacles, the proposed bio-inspired soft gripper provides the mechanical foundation for multi-mode grasping. 
With the goal of achieving human-like stable long-duration grasp maintenance, the system adopts a mechanism–AI co-design framework that integrates multi-mode grasping structures, multimodal perception (vision, temperature, and learning-based force estimation), and physics-informed force regulation.
}
    \label{fig:overview}
\end{figure}

In contrast to rigid dexterous hands that depend on high-fidelity modeling and high-bandwidth closed-loop control, soft robotic grippers embed compliance directly into their mechanical structure, approximating the physical adaptivity of biological tissues with substantially reduced system complexity.
This intrinsic compliance significantly improves contact robustness when interacting with unknown or irregular objects \cite{trivedi2008soft,rus2015design}.
Representative examples include pneumatically actuated fingers \cite{9380478}, bio-inspired Fin Ray architectures \cite{yang2017bioinspired,chen2019bio, guo2026construction}, tendon-driven compliant structures \cite{zhang2022bioinspired}, and soft polyhedral networks \cite{liu2024proprioceptive}, all of which demonstrate that structural compliance can effectively reduce reliance on precise modeling and planning \cite{wang2023online, dong2018real, wang2025sa}. However, the same compliance that enables gentle contact also reduces load capacity and force stability during holding.

To mitigate this trade-off, a variety of variable-stiffness mechanisms have been introduced to enhance the mechanical adaptability of soft grippers across different manipulation stages.
Existing approaches include mechanically induced stiffening strategies based on granular or layer jamming \cite{9730080}, kinematic reconfiguration \cite{bastien2023variable}, as well as material-based routes that exploit phase transitions  or field-responsive effects, such as electrorheological \cite{hao2001electrorheological}, magnetorheological \cite{de2011magnetorheological}, low-melting-point alloys \cite{zhao2016development}, shape memory alloys \cite{11000132}, and shape memory polymers (SMPs).
Among these options, SMPs have emerged as a representative material choice for variable-stiffness soft grippers due to their large, reversible modulus variation and favorable structural integrability.  \cite{behl2007shape, luo2024recent,firouzeh2017stiffness, guo2025enabling}.
Through thermal activation, SMP-based soft fingers can transition between highly compliant configurations for contact acquisition and stiff configurations for load-bearing, partially emulating the phase-dependent stiffness regulation observed in human musculature.
Compared with alternative approaches, SMP-based stiffness modulation avoids reliance on strong external fields, rigid inclusions, or additional mechanical reconfiguration, making it particularly suitable for achieving continuous and spatially distributed stiffness tuning within deformable soft structures.

However, stiffness modulation alone remains insufficient to ensure stable force output over long durations.
Polymer-based materials inherently exhibit viscoelastic behavior, which leads to pronounced \textbf{\emph{Stress Relaxation}} during sustained contact: \textbf{even when geometry and actuation remain unchanged, interaction forces decay over time} 
\cite{gutierrez2014engineering,shintake2018soft}.
In thermally actuated variable-stiffness systems, this challenge is further compounded, as temperature not only determines instantaneous stiffness but also directly governs the relaxation dynamics of the material. To compensate for such long-term force decay, the system must continuously perceive or reconstruct the evolving interaction force during holding without compromising the compliance and large-deformation capability of the soft gripper.
Most existing perception approaches can be organized by their observation mechanism into direct measurement and indirect inference methods.
Direct measurement methods embed force, strain, pressure, or tactile sensing units in the soft body or contact interface
\cite{yang2023self,9999176,ji2021gradient,liu20201,dagdeviren2014conformable,tomo2017covering,massari2019tactile}.
They provide local feedback, but the resulting sensor--body coupling may affect integration and compliance under large deformation.
Inference-based methods are less intrusive and estimate force from deformation, pose, structural responses \cite{wang2025flexible,li2023jamtac}, optical waveguide modulation \cite{kim2020heterogeneous}, or acoustic sensing \cite{wall2023passive}.
However, their structural-response--force mapping can drift in viscoelastic soft structures during long-term interaction if relaxation is not explicitly modeled.
Although recent work has begun to model viscoelasticity temporally \cite{liu2024proprioceptive}, long-horizon closed-loop force compensation for variable-stiffness soft grippers remains largely unexplored.

From this perspective, achieving stable long-duration grasping requires the coordinated integration of at least four interdependent elements:
\textbf{compliant structures for reliable contact acquisition},
\textbf{tunable stiffness for phase-dependent mechanical adaptation},
\textbf{continuous multimodal perception for observing interaction states}, as well as
\textbf{explicit modeling and compensation of time-dependent force evolution}.
Existing systems typically address only subsets of these capabilities, and a soft grasping platform that systematically integrates all four elements toward the goal of stable grasp maintenance is still lacking.

To address this gap, this paper presents a variable-stiffness soft gripper together with a relaxation-aware multimodal sensing and force-regulation framework (Fig.~\ref{fig:overview}). The main contributions are highlighted below:

\textbf{First}, a bio-inspired soft finger is introduced that combines passive compliance, thermally tunable stiffness, and switchable adhesion within a unified structure, enabling multimode grasping.
It continuously senses and modulates interaction conditions across pre-contact, contact establishment, and holding, rather than relying solely on instantaneous contact feedback.

\textbf{Second}, an integrated multimodal perception architecture is developed to maintain non-invasive interaction observability under large deformation.
Vision provides proximity cues and proprioceptive deformation tracking, infrared thermography captures the temperature field, and a model-based inference module fuses these signals to reconstruct contact force.

\textbf{Third}, a temperature-coupled reduced viscoelastic force representation is formulated to separate the interaction force into instantaneous and relaxation-induced components.
A physics-constrained learning framework then reconstructs force evolution and enables long-horizon compensation during holding.

\textbf{Fourth}, experiments demonstrate stable long-duration force regulation in soft robotic grasping enabled by the coordinated integration of stiffness modulation, multimodal perception, and relaxation-aware learning. These results support a mechanism--AI co-design paradigm for soft manipulation under coupled thermo-mechanical and viscoelastic dynamics.

\section{Gripper Design}

\begin{figure}[t]
    \centering
    \includegraphics[width=1\linewidth]{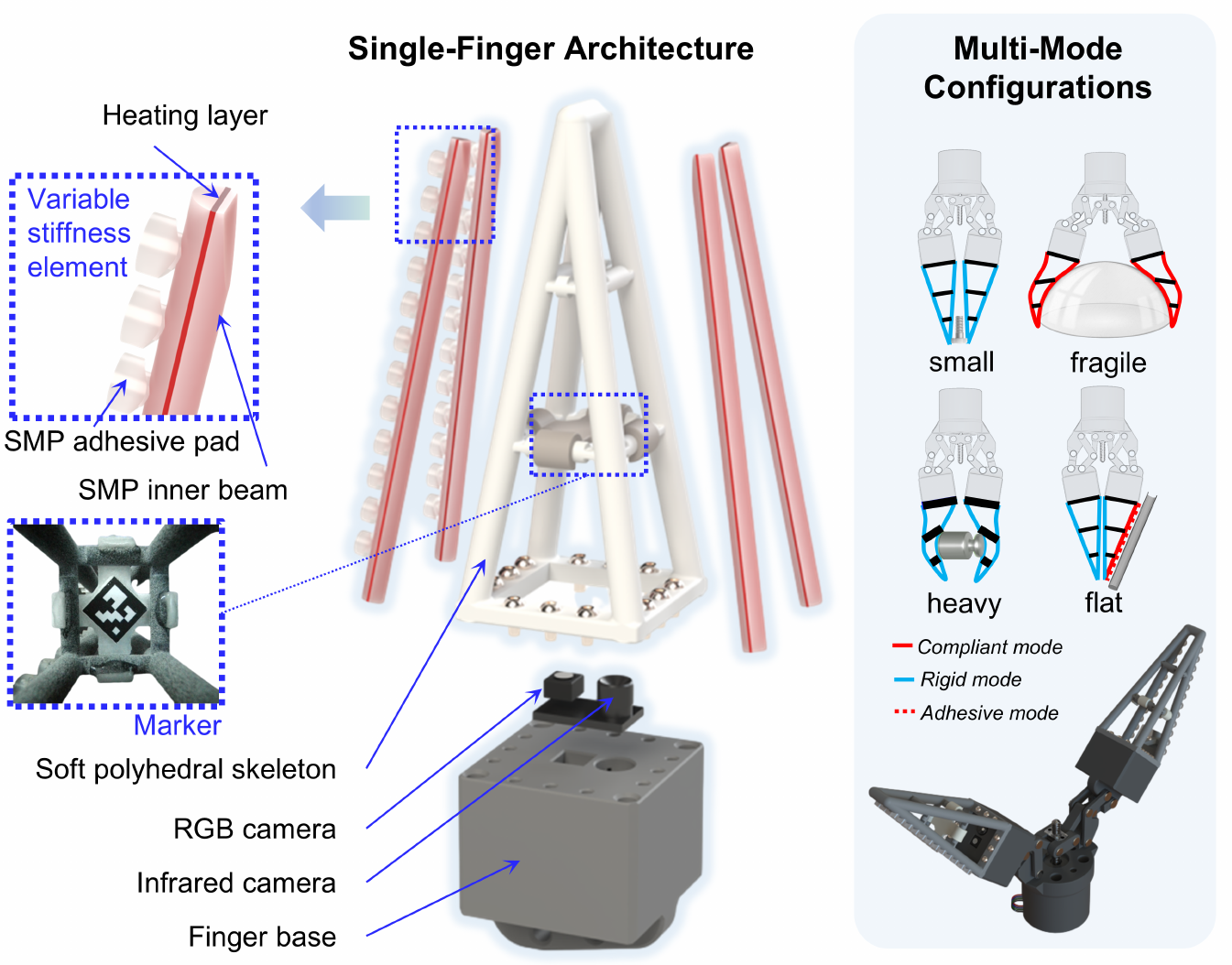}
\caption{
Overview of the single-finger architecture and multi-mode grasping configurations.
The exploded view illustrates the compliant skeletal architecture of a single finger, highlighting the embedded thermo-responsive SMP beams together with integrated heating and sensing components.
}

    \label{fig:finger}
\end{figure}

\begin{figure}[t]
    \centering
    \includegraphics[width=1\linewidth]{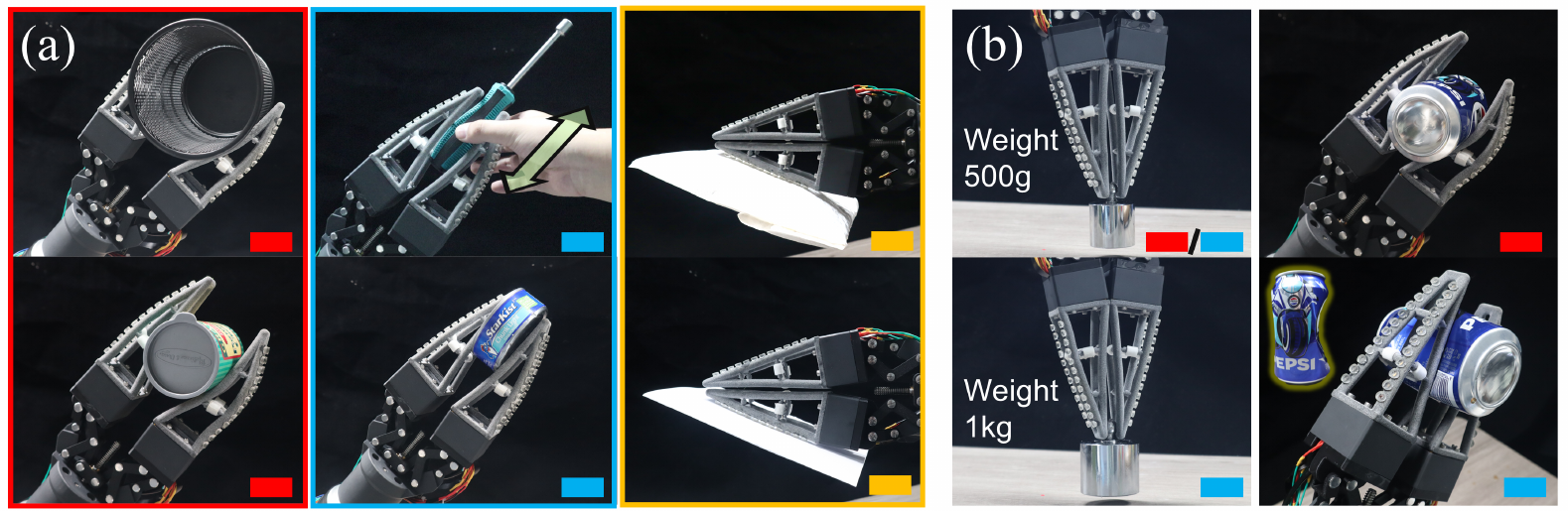}
    \caption{
    Multi-mode grasping enabled by stiffness modulation and switchable adhesion.
    (a) Three grasping modes: compliant grasp (red) for conformal contact, stiff grasp (blue) for load-bearing holding and disturbance rejection (the screwdriver remains secured under manual perturbation), and adhesive grasp (orange) for attachment-assisted grasping.
    (b) Stiffness-dependent trade-offs: the stiff mode lifts a 1kg weight by fingertip pinch, whereas the compliant mode cannot; conversely, the compliant mode grasps an empty aluminum can without visible deformation, while the stiff mode crushes it.
    }
    \label{fig:modes}
\end{figure}

The proposed soft gripper, shown in Fig.~\ref{fig:finger}, consists of two identical soft fingers. It integrates thermally tunable stiffness with base-mounted vision and thermal sensing in a compact system. The system senses and regulates interaction conditions throughout approach, contact formation, and holding, rather than relying on a single instantaneous contact measurement.

Each finger adopts a network-like open skeletal architecture \cite{liu2024proprioceptive}, where axially aligned load-bearing beams are connected by compliant joints. This morphology preserves mechanical integrity while enabling large passive deformation and omnidirectional geometric adaptation, so the finger conforms to object shape without relying on precise kinematic alignment. The open cavity further provides an unobstructed sensing corridor under large deformation, which supports non-invasive proprioceptive observation.

Thermally tunable stiffness is realized by embedding temperature-responsive SMP beams within the skeleton. Temperature regulation switches the finger between a compliant mode for conformal contact and a stiff mode for load-bearing holding, without changing the external geometry or requiring mechanical reconfiguration. Switchable adhesion is enabled by thermally activated pads on the finger surface, allowing attachment to planar or low-curvature surfaces to complement force-closure grasping. Figure~\ref{fig:modes} illustrates three grasping modes (compliant, stiff, adhesive) and shows that stiffness switching trades load-bearing and disturbance rejection against gentle, non-damaging grasping.

To preserve compliance, sensing modules are mounted on the gripper base rather than embedded within the soft structure. An RGB camera tracks a single ArUco marker to capture global configuration changes, while an infrared thermal camera monitors the temperature field associated with stiffness modulation. With fixed extrinsics, the two sensors provide synchronized observation of deformation and thermal state. This layout supports three perceptual functions: visual proximity sensing during approach, vision-based force inference from marker motion, and non-contact temperature sensing via infrared thermography. 

Details of the thermo-responsive SMP material properties (including dynamic mechanical analysis) and the finger fabrication process are provided in the Supplementary Materials.

\section{Physics-informed Force Perception and Relaxation Compensation}

This section presents a physics-informed force perception and relaxation compensation framework for variable-stiffness soft grippers under thermally induced viscoelasticity.
A reduced force representation is derived to separate instantaneous elastic response from time-dependent relaxation effects.
Based on this formulation, a physics-aligned learning architecture reconstructs interaction forces from multimodal observations and compensates relaxation-induced force decay during prolonged grasping.

\begin{figure*}[t]
    \centering
    \includegraphics[width=1\linewidth]{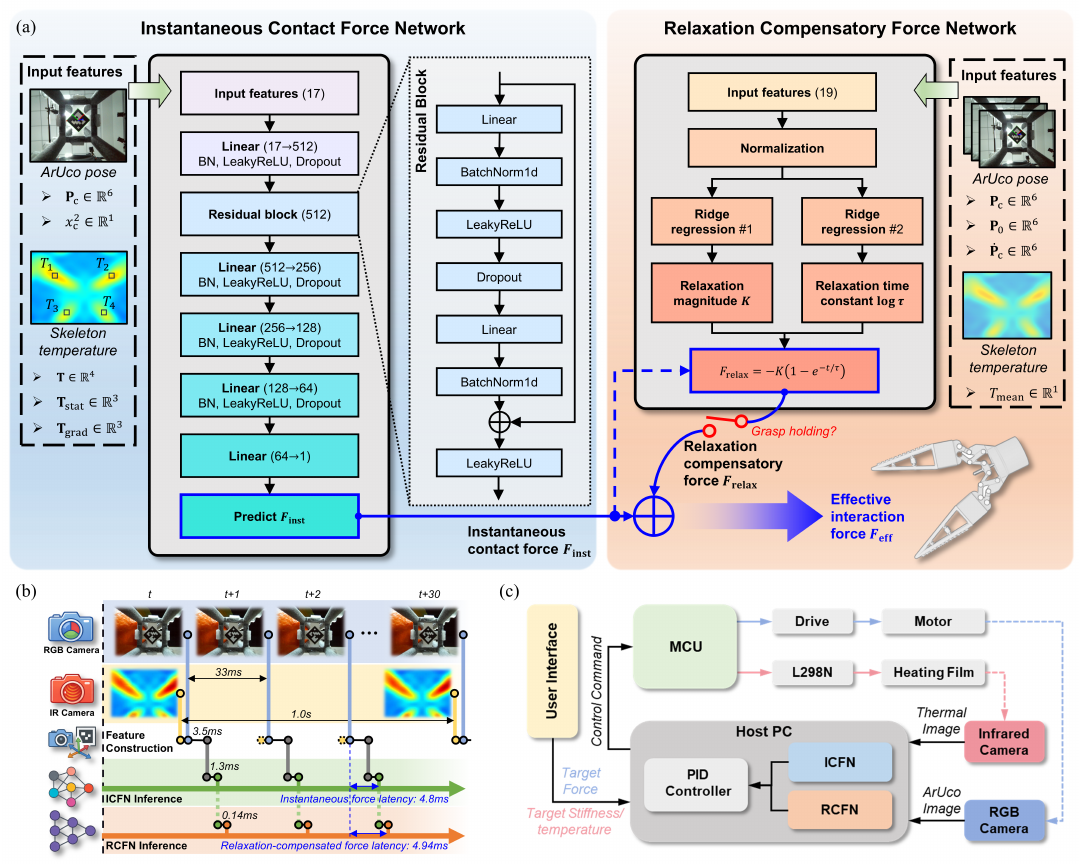}
    \caption{
    Physics-aligned learning and force-regulation pipeline.
    (a) ICFN estimates the instantaneous contact force from pose and thermal features, while RCFN predicts temperature-dependent relaxation parameters for holding-phase compensation.
    (b) Online multimodal data flow and measured inference timing for instantaneous and relaxation-compensated force estimation.
    (c) Closed-loop system architecture for force regulation and thermal stiffness control.
    }
    \label{fig:network}
\end{figure*}

\subsection{Viscoelastic Force Modeling Under Displacement-Controlled Grasping}

The interaction force generated by polymer-based soft fingers is governed by temperature-dependent viscoelasticity.
To capture force evolution during prolonged holding, we model the thermomechanical response using a generalized Maxwell--Wiechert formulation.

The model consists of an equilibrium elastic spring with stiffness $k_e(T)$, where $T$ denotes temperature, in parallel with $N$ Maxwell branches.
Each Maxwell branch comprises a spring $k_i(T)$ and a dashpot $\eta_i(T)$ in series.
All parameters are explicitly temperature dependent, reflecting thermally induced modulus softening and accelerated molecular mobility.

Under displacement-controlled contact with displacement $X(t)$, the total interaction force is
\begin{equation}
F(t)=k_e(T)X(t)+\sum_{i=1}^{N}F_i(t),
\label{eq:wiechert_force}
\end{equation}
where $F_i(t)$ denotes the internal force of the $i$-th Maxwell branch.
For each branch,
\begin{equation}
\tau_i(T)\dot{F}_i(t)+F_i(t)=k_i(T)\tau_i(T)\dot{X}(t),
\label{eq:maxwell_branch}
\end{equation}
with relaxation time constant $\tau_i(T)=\eta_i(T)/k_i(T)$.

After contact establishment, the holding phase can be approximated as a step displacement input $X(t)=X_0$.
Under this condition,
\begin{equation}
F_i(t)=k_i(T)X_0 e^{-t/\tau_i(T)},
\end{equation}
yielding the total force
\begin{equation}
F(t)=k_e(T)X_0+\sum_{i=1}^{N}k_i(T)X_0 e^{-t/\tau_i(T)}.
\label{eq:force_full}
\end{equation}

\subsection{Reduced Temperature-Coupled Relaxation Representation}

We define the instantaneous contact force at the onset of holding as
\begin{equation}
F_{\mathrm{inst}} \triangleq F(0)
= X_0\!\left(k_e(T)+\sum_{i=1}^{N}k_i(T)\right).
\label{eq:inst_force}
\end{equation}

Rearranging Eq.~\eqref{eq:force_full} gives
\begin{equation}
F(t)=F_{\mathrm{inst}}
- X_0\sum_{i=1}^{N}k_i(T)\!\left(1-e^{-t/\tau_i(T)}\right),
\label{eq:force_loss}
\end{equation}
where the second term represents the cumulative viscoelastic relaxation loss.

Over task-relevant holding horizons, fast relaxation modes decay shortly after contact establishment, while slow modes dominate the long-term force evolution.
Accordingly, we approximate the aggregate relaxation behavior using a dominant effective mode,
\begin{equation}
\sum_{i=1}^{N}k_i(T)\!\left(1-e^{-t/\tau_i(T)}\right)
\approx
K(T)\!\left(1-e^{-t/\tau(T)}\right),
\label{eq:dominant_mode}
\end{equation}
where $K(T)$ and $\tau(T)$ denote the effective relaxation magnitude and time constant.

Substituting Eq.~\eqref{eq:dominant_mode} into Eq.~\eqref{eq:force_loss}, the force evolution during holding is expressed as
\begin{equation}
F(t)=F_{\mathrm{inst}}
- K(T)\!\left(1-e^{-t/\tau(T)}\right).
\label{eq:reduced_force}
\end{equation}

This reduced representation explicitly separates the instantaneous elastic response from subsequent relaxation-driven force decay, while preserving the essential temperature dependence of soft-material dynamics.

\subsection{Physics-Aligned Learning and Force-Regulation Pipeline}

Rather than identifying the full set of viscoelastic parameters, we directly learn the reduced quantities in Eq.~\eqref{eq:reduced_force}.
As shown in Fig.~\ref{fig:network}a, two complementary modules are constructed.

\emph{Instantaneous Contact Force Network (ICFN)} estimates $F_{\mathrm{inst}}$ from multimodal observations at contact establishment.
The input features include the ArUco-based pose, squared marker translation, beam temperatures, and engineered thermal features such as temperature statistics and spatial gradients.
These features jointly encode finger deformation, thermal softening, and spatially non-uniform stiffness distribution.
\emph{Relaxation Compensation Force Network (RCFN)} predicts the effective relaxation parameters $(K,\tau)$ from the current thermal and kinematic context during holding.
Its input includes the current temperature, the contact-onset pose, the current pose, and the pose velocity.
In implementation, RCFN is realized by two lightweight Ridge regressors that predict the relaxation magnitude and the logarithm of the relaxation time constant.

The reconstructed interaction force is given by
\begin{equation}
\hat{F}(t)=\hat{F}_{\mathrm{inst}}
- \hat{K}\!\left(1-e^{-t/\hat{\tau}}\right),
\label{eq:force_recon}
\end{equation}
enabling explicit compensation of temperature-dependent viscoelastic relaxation in closed-loop force regulation.
This formulation separates the instantaneous contact response from relaxation-induced force decay, making long-horizon force drift observable, learnable, and compensable online. Detailed feature definitions, hyperparameters, and the ablation study are provided in the Supplementary Materials.

Figure~\ref{fig:network}b summarizes the online multimodal data flow and measured inference timing.
The two sensing streams have different update rates and roles.
RGB images are updated at approximately 30 Hz and are used for ArUco-based pose and deformation tracking, while infrared images are updated at approximately 1 Hz and provide beam-level thermal feedback that reflects the thermal state of the fingers.
At contact establishment, these observations are converted into compact multimodal features and fed into the ICFN to estimate the instantaneous interaction force $F_{\mathrm{inst}}$.
In our implementation, ArUco pose extraction takes approximately 3.5 ms, feature construction is lightweight, and ICFN inference takes approximately 1.3 ms, yielding an instantaneous force latency of approximately 4.8 ms.
During holding, the RCFN predicts the effective relaxation parameters from the current thermal state and kinematic context.
The steady-state RCFN inference adds approximately 0.14 ms, resulting in a relaxation-compensated force latency of approximately 4.94 ms.

Figure~\ref{fig:network}c shows the corresponding closed-loop system architecture.
The host PC serves as the central processing unit for multimodal perception and force regulation.
It fuses RGB and infrared observations, reconstructs the interaction force through the ICFN and RCFN, and sends the reconstructed force to the PID controller.
The PID controller outputs velocity commands through the MCU to drive the gripper motor and maintain the desired grasp force.
Meanwhile, a temperature control loop regulates the heating film through an L298N driver, enabling continuous stiffness adjustment via thermal actuation.
Target force and target stiffness or temperature are specified through the user interface, supporting phase-dependent interaction strategies across contact formation and holding.
Together, this perception--learning--control loop enables the system to compensate relaxation-induced force decay online rather than relying only on instantaneous contact feedback.

\section{Experiments}

This section presents a series of experiments to evaluate the performance of the proposed variable-stiffness soft gripper, with a particular focus on multimodal force perception accuracy and long-duration grasp stability under thermally coupled viscoelastic effects.

\subsection{Temperature-Dependent Stiffness Characterization}

\begin{figure}
    \centering
    \includegraphics[width=1\linewidth]{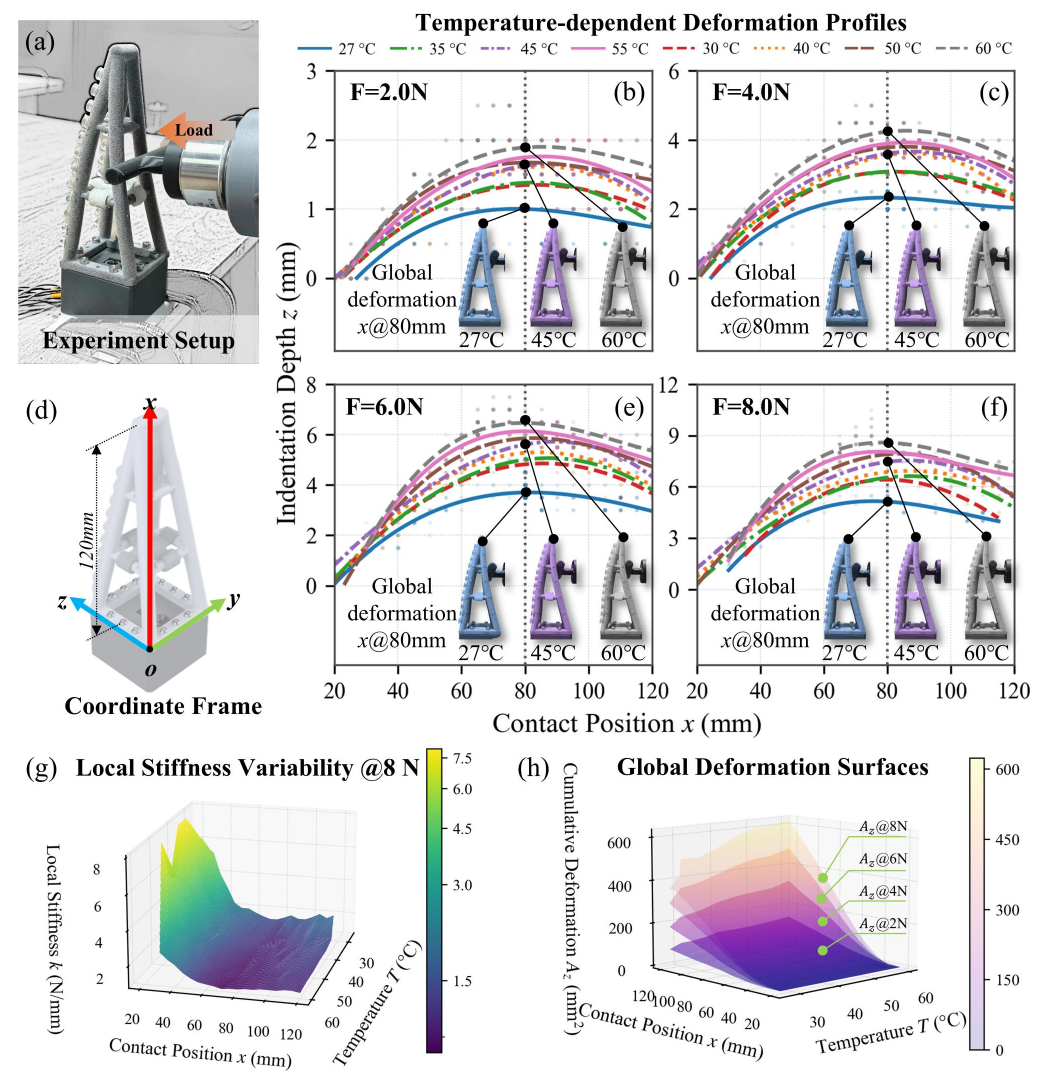}
    \caption{
    Temperature-dependent deformation characterization of the soft finger.
    (a) Experiment setup.
    (d) Definition of the local finger coordinate frame.
(b, c, e, f) Deformation profiles $z(x)$ under different forces and temperatures.
(g) Local stiffness variation at 8N as a function of contact position and temperature.
(h) Global deformation surfaces $A_z$ under different loads, illustrating coupled effects of contact position and temperature.
    }
    \label{fig:deform}
\end{figure}

We characterize the temperature-dependent stiffness of the proposed SMP-integrated soft finger to quantify how temperature, contact location, and normal load jointly shape contact-induced deformation, providing a physical basis for subsequent force modeling.

As shown in Fig.~\ref{fig:deform}a, the finger is rigidly fixed while a UR5e robot applies controlled normal indentation; a local frame is defined with $x$ along the finger and $z$ as the normal deformation direction (Fig.~\ref{fig:deform}d). Deformation profiles $z(x)$ are measured under 2--8N and 27--60$^\circ$C, spanning the glassy--rubbery transition (Fig.~\ref{fig:deform}b,c,e,f). Deformation increases monotonically with temperature across all loads, indicating pronounced thermomechanical softening.

Local stiffness is evaluated at $x=80$mm as $k_{80}=F/z_{80}$. Increasing temperature from 27$^\circ$C to 60$^\circ$C yields a $1.6\times$--$1.7\times$ increase in deformation, corresponding to a 35\%--45\% reduction in $k_{80}$ across force levels. Global deformation, quantified by $A_z=\int z(x)\,dx$, increases with both temperature and load (e.g., at 8N, 361mm$^2$ to 626mm$^2$, +73\%), suggesting system-level softening rather than a localized effect. The coupled spatial--thermal patterns are summarized in Fig.~\ref{fig:deform}g,h, visualizing the joint influence of contact position, temperature, and load on local stiffness and global deformation.

These results establish the finger's temperature-dependent stiffness characteristics and motivate the subsequent physics-informed modeling of instantaneous contact force and temperature-dependent viscoelastic relaxation.

\subsection{Force Perception and Relaxation Reconstruction Accuracy}

We evaluate the proposed framework in two aspects: ICFN and RCFN, using held-out test data and targeted repeatability tests.

\begin{figure}
    \centering
    \includegraphics[width=1\linewidth]{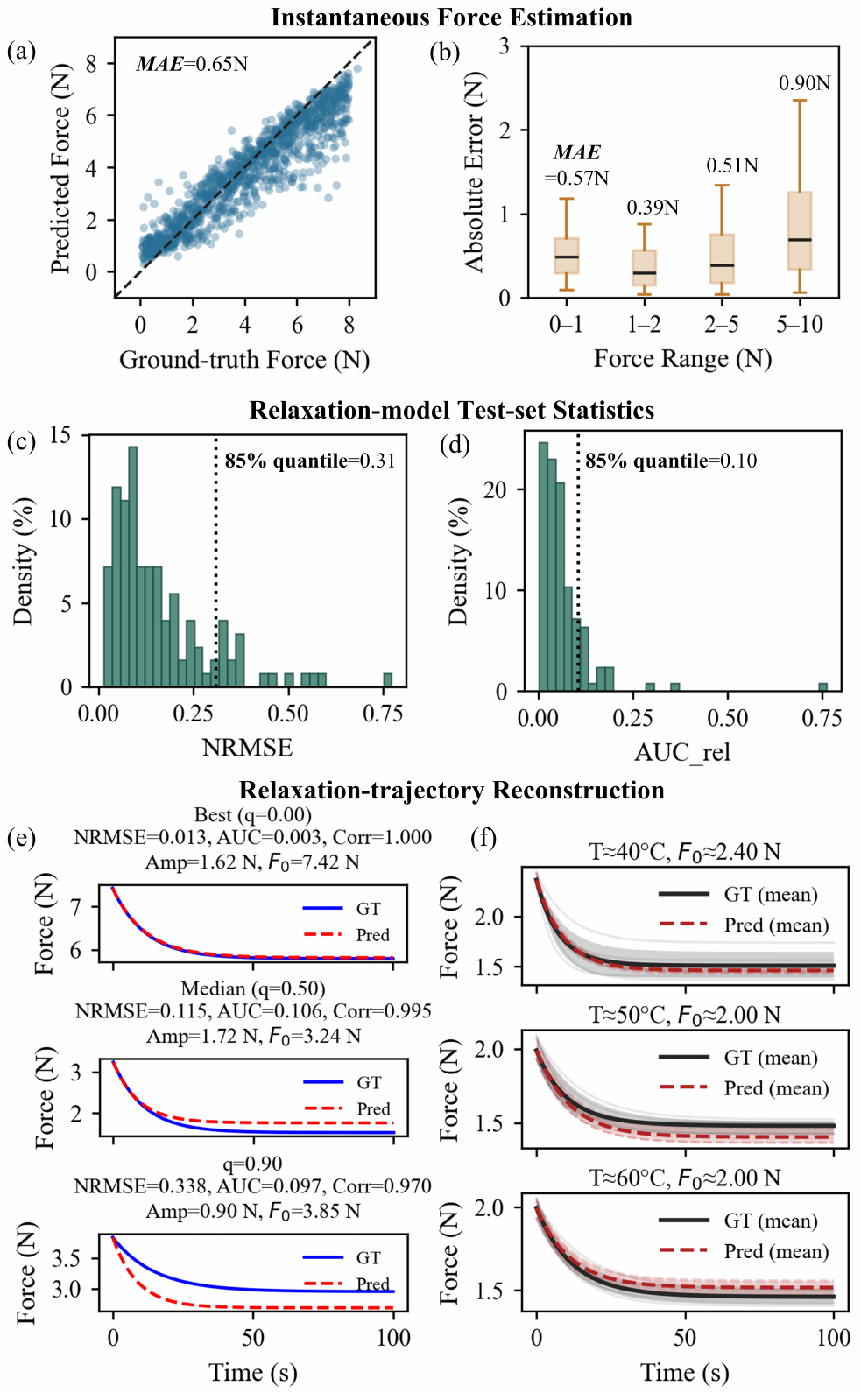}
    \caption{
    Force perception accuracy and relaxation modeling results.
    (a) ICFN parity on the test set.
    (b) ICFN absolute error across force ranges.
    (c--d) RCFN test-set distributions of NRMSE and AUC$_{\mathrm{rel}}$.
    (e) RCFN reconstructions on representative random samples at three quantiles ($q=0.0,0.5,0.9$).
    (f) Repeatability test over six trials at similar $F_0$ under different temperatures.
    }
    \label{fig:acc}
\end{figure}

\paragraph{Metrics}
We report pointwise force errors for ICFN and curve-level metrics for RCFN.
For relaxation curves, NRMSE measures normalized waveform deviation,
\[
\mathrm{NRMSE}
=
\frac{\sqrt{\frac{1}{T}\int_{0}^{T}\!\left(F_{\mathrm{pred}}(t)-F_{\mathrm{gt}}(t)\right)^2 dt}}
{\max(F_{\mathrm{gt}})-\min(F_{\mathrm{gt}})},
\]
and AUC$_{\mathrm{rel}}$ quantifies relative cumulative drift,
\[
\mathrm{AUC}_{\mathrm{rel}}
=
\frac{\left|\int_{0}^{T}\!\left(F_{\mathrm{pred}}(t)-F_{\mathrm{gt}}(t)\right)dt\right|}
{\int_{0}^{T}\!F_{\mathrm{gt}}(t)dt}.
\]
We also use the relaxation amplitude $\mathrm{Amp}=F_0-F_{\mathrm{end}}$ to characterize relaxation strength.

\paragraph{Datasets}
ICFN is trained on 8,258 samples spanning 25--60$^\circ$C and forces up to 8N.
RCFN is trained on 810 relaxation sequences (100s) under fixed contact, covering four contact locations and multiple temperature--force conditions.
All data are split into 70\%/15\%/15\% training/validation/test sets.

\paragraph{Instantaneous force estimation (ICFN)}
ICFN predictions match ground truth across stiffness regimes (Fig.~\ref{fig:acc}a).
On the test set, ICFN achieves an MAE of 0.651 N, corresponding to a full-scale error of 6.51\% over the 0--10 N force range, with an RMSE of 0.886 N and $R^2=0.848$.
Errors remain bounded across force ranges (Fig.~\ref{fig:acc}b), with median absolute errors of
0.489N (0--1N), 0.297N (1--2N), 0.393N (2--5N), and 0.695N (5--10N),
indicating stable estimation from weak to high-load contacts. 

\paragraph{Relaxation reconstruction (RCFN)}
RCFN predicts the effective relaxation parameters and reconstructs the full holding-force trajectory.
On the test set, NRMSE attains mean/median 0.160/0.118 and AUC$_{\mathrm{rel}}$ remains low at 0.065/0.049 (Fig.~\ref{fig:acc}c,d),
with a mean curve-level correlation of 0.990.
Fig.~\ref{fig:acc}e further visualizes reconstruction quality using three representative random samples selected at error quantiles ($q=0.0,0.5,0.9$),
showing that RCFN captures both relaxation magnitude and temporal evolution.

To assess repeatability, we group trials by temperature with matched initial forces and evaluate six randomly sampled sequences per group (Fig.~\ref{fig:acc}f).
The relaxation reconstruction remains consistent across repeated contacts. At $\sim$40$^\circ$C, the model achieves $\mathrm{NRMSE}=0.091\pm0.040$ and $\mathrm{AUC}_{\mathrm{rel}}=0.029\pm0.018$ with a high correlation (Corr$=0.997\pm0.005$) under low end-force variability in the ground truth (GT std at $t{\mathrm{end}}=0.041$N). Similar consistency is observed at $\sim$50$^\circ$C ($\mathrm{NRMSE}=0.143\pm0.068$, $\mathrm{AUC}_{\mathrm{rel}}=0.042\pm0.015$, Corr$=0.991\pm0.010$; GT std 0.040N). At $\sim$60$^\circ$C, although the spread increases ($\mathrm{NRMSE}=0.140\pm0.099$, $\mathrm{AUC}_{\mathrm{rel}}=0.067\pm0.042$; GT std 0.121N), the correlation remains high (Corr$=0.991\pm0.013$), indicating that the model reliably captures the trajectory shape of relaxation across repeats. 

Overall, these results support accurate instantaneous force perception, reliable relaxation reconstruction across temperatures and repeated contacts.

\subsection{Holding Force Measurement and Relaxation-Aware Force Control}

\begin{figure*}[ht]
    \centering
    \includegraphics[width=1\linewidth]{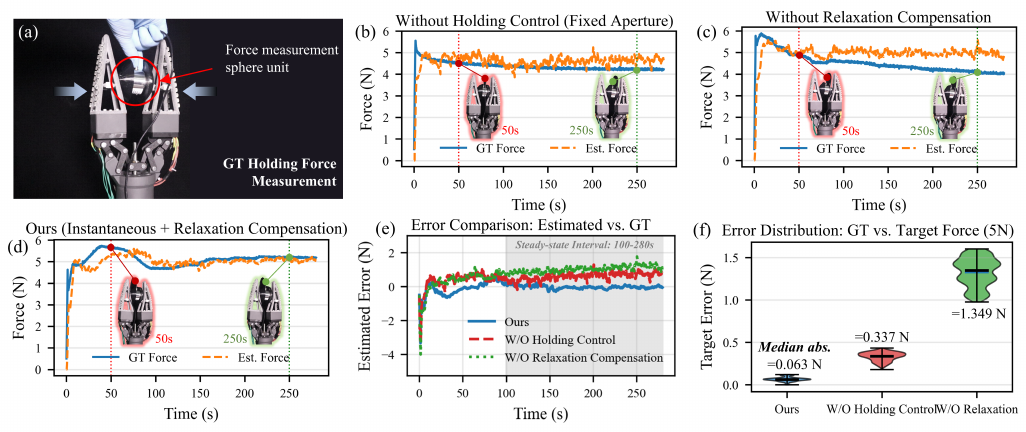}
    \caption{
Holding-force evaluation and relaxation-aware force regulation.
(a) Experimental setup for ground-truth holding-force measurement using a spherical force measurement unit.
(b--d) Long-duration holding-force responses under three configurations: without holding control, without relaxation compensation, and the proposed ICFN+RCFN method. Blue curves denote ground-truth force and orange dashed curves denote estimated force.
(e) Estimated-force error over time. The shaded region denotes the steady-state interval used for quantitative evaluation.
(f) Distribution of target-force error with respect to the 5 N target. 
}
    \label{fig:sphere}
\end{figure*}

We evaluate the proposed relaxation-aware force perception and holding strategy during prolonged grasping, with the objective of assessing force estimation accuracy and holding stability under temperature-dependent viscoelastic relaxation.

As shown in Fig.~\ref{fig:sphere}a, the gripper grasps a spherical force measurement unit equipped with an internal pressure sensor that provides ground-truth normal force.
After contact establishment, the grasp is maintained for approximately 280s without enforcing precise object positioning.
Three configurations are compared:
(i) a fixed-aperture baseline without holding regulation (\emph{Without Holding Control}),
(ii) an instantaneous-estimation baseline using ICFN only (\emph{Without Relaxation Compensation}),
and (iii) the proposed relaxation-aware method integrating ICFN and RCFN (\emph{Ours}).

Results are shown in Fig.~\ref{fig:sphere}b--d.
With a fixed aperture, the holding force decays steadily due to viscoelastic relaxation and cannot be compensated, leading to growing deviation from the target (Fig.~\ref{fig:sphere}b).
Using instantaneous force estimation alone partially stabilizes the estimated force, but systematically overestimates the true interaction force as relaxation progresses (Fig.~\ref{fig:sphere}c).
In contrast, the proposed method accurately tracks relaxation-induced force evolution and maintains stable regulation around the 5N target throughout the entire holding duration (Fig.~\ref{fig:sphere}d).

Quantitative error statistics over the steady-state interval (100--280s) are summarized in Fig.~\ref{fig:sphere}e,f.
The proposed method exhibits negligible bias (mean signed error $-0.036$N, median $-0.041$N) and low dispersion (RMSE 0.102N).
By comparison, the fixed-aperture baseline shows a larger bias and error (mean 0.606N, RMSE 0.643N),
while the instantaneous-only baseline suffers from severe overestimation (mean 1.019N, RMSE 1.035N).
Relative to the 5N target, the proposed approach achieves a mean absolute error of 0.066 N, corresponding to only 1.32\% of the target force, with a narrow interquartile range of [0.055, 0.076] N. In comparison, the fixed-aperture baseline has an MAE of 0.333 N, corresponding to 6.66\% of the target force, while the instantaneous-only baseline has an MAE of 1.327 N, corresponding to 26.54\% of the target force. Thus, the proposed method reduces MAE by approximately 80.2\% compared with the baseline without holding control and by 95.0\% compared with the baseline without relaxation compensation.

These results demonstrate that explicitly modeling and compensating for temperature-dependent viscoelastic relaxation is critical for accurate force perception and stable long-duration grasping in soft robotic systems.

\subsection{Real-Time Force Perception for Stiffness-dependent Stable Grasping}

\begin{figure*}[ht]
    \centering
    \includegraphics[width=1\linewidth]{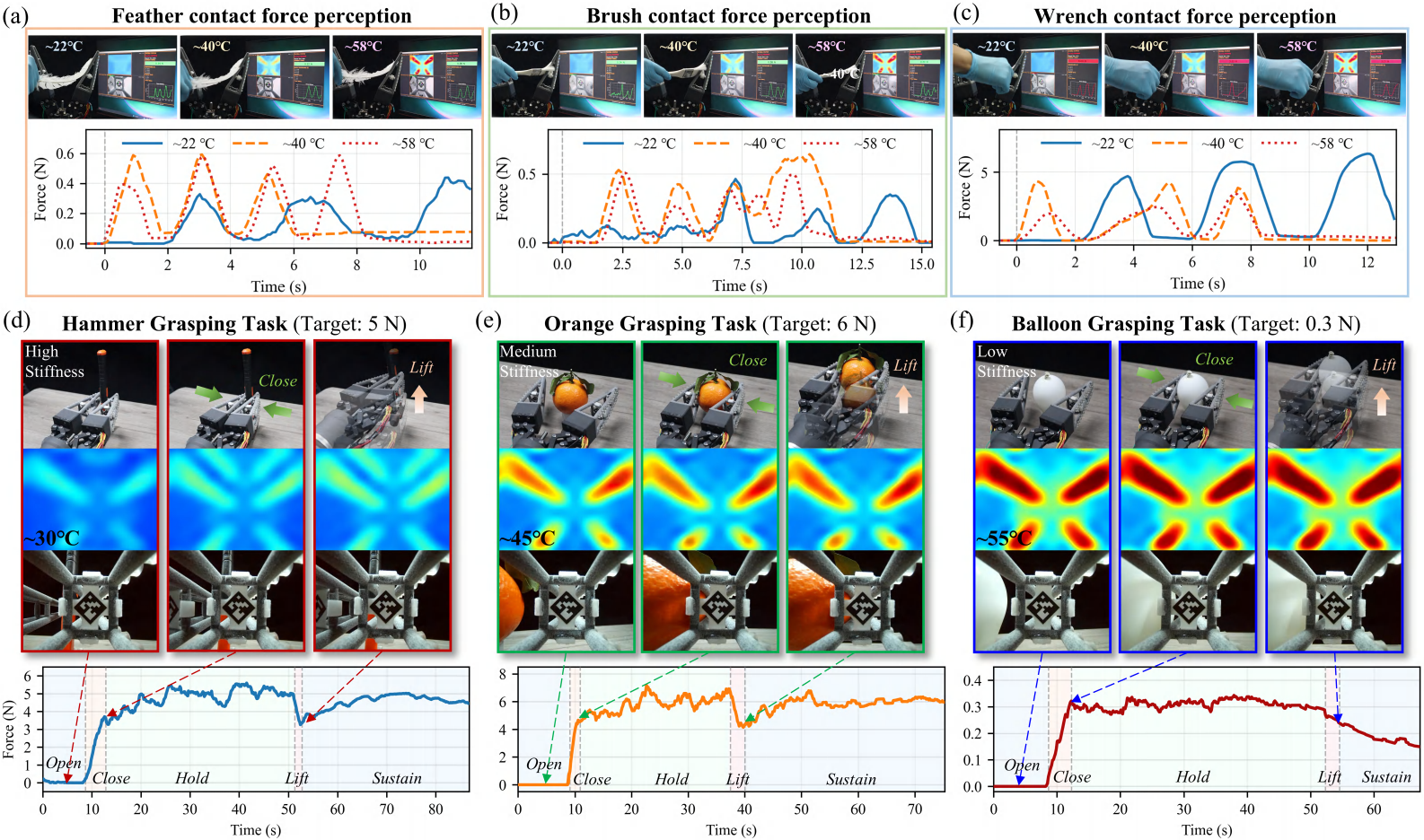}
\caption{
Real-time force perception under stiffness modulation and closed-loop grasping.
(a--c) Single-finger contact-force perception with a feather, brush, and wrench under three temperature-controlled stiffness states: high stiffness at approximately 22$^\circ$C, medium stiffness at approximately 40$^\circ$C, and low stiffness at approximately 58$^\circ$C.
(d--f) Closed-loop grasping tasks with stiffness-dependent strategies: hammer grasping at high stiffness with a 5 N target, orange grasping at medium stiffness with a 6 N target, and balloon grasping at low stiffness with a 0.3 N target. The stages Open, Close, Hold, Lift, and Sustain indicate the phase-dependent grasping process. Higher temperature corresponds to lower finger stiffness.
}
    \label{fig:grasp}
\end{figure*}

Figure~\ref{fig:grasp} illustrates how variable stiffness modulation interacts with real-time force perception and shapes physical interaction behavior.

Single-finger contact experiments (Fig.~\ref{fig:grasp}a--c) reveal pronounced stiffness-dependent force responses.
At lower temperatures (higher stiffness), contact induces force transients with larger amplitudes and steeper temporal gradients,
whereas more compliant configurations produce smoother responses with reduced force magnitude.
Despite these variations, instantaneous force estimates remain stable and consistent across all stiffness states, indicating that ICFN reliably infers contact force independently of stiffness-dependent deformation patterns.

Weak-contact experiments with a feather and a brush further demonstrate the sensitivity of the perception framework.
Even brief and lightweight contacts generate clearly detectable force signals, which are amplified in compliant configurations due to increased structural deformability.
These results highlight a stiffness-dependent sensing trade-off:
softer configurations enhance sensitivity to weak interactions, while stiffer configurations favor the generation and maintenance of larger forces.

Closed-loop grasping experiments (Fig.~\ref{fig:grasp}d--f) demonstrate how stiffness selection and force regulation jointly enable task-appropriate manipulation.
A high-stiffness configuration sustains a 5N target force during hammer lifting with minimal drift and high load capacity.
An intermediate stiffness enables stable force regulation when grasping an orange while avoiding excessive contact pressure.
In a highly compliant configuration, the system reliably regulates sub-Newton forces to grasp a balloon without visible deformation.

Across all tasks, the estimated force remains tightly regulated around the desired target throughout approach, closure, holding, and lifting phases.
Together, these results show that stiffness modulation acts as a structural prior that constrains feasible force ranges and interaction dynamics,
while real-time force perception provides the feedback necessary to achieve stable force-closure grasping across diverse stiffness regimes.

\subsection{Integrated Demonstration of Variable Stiffness and Multimodal Perception}

\begin{figure*}[ht]
    \centering
    \includegraphics[width=1\linewidth]{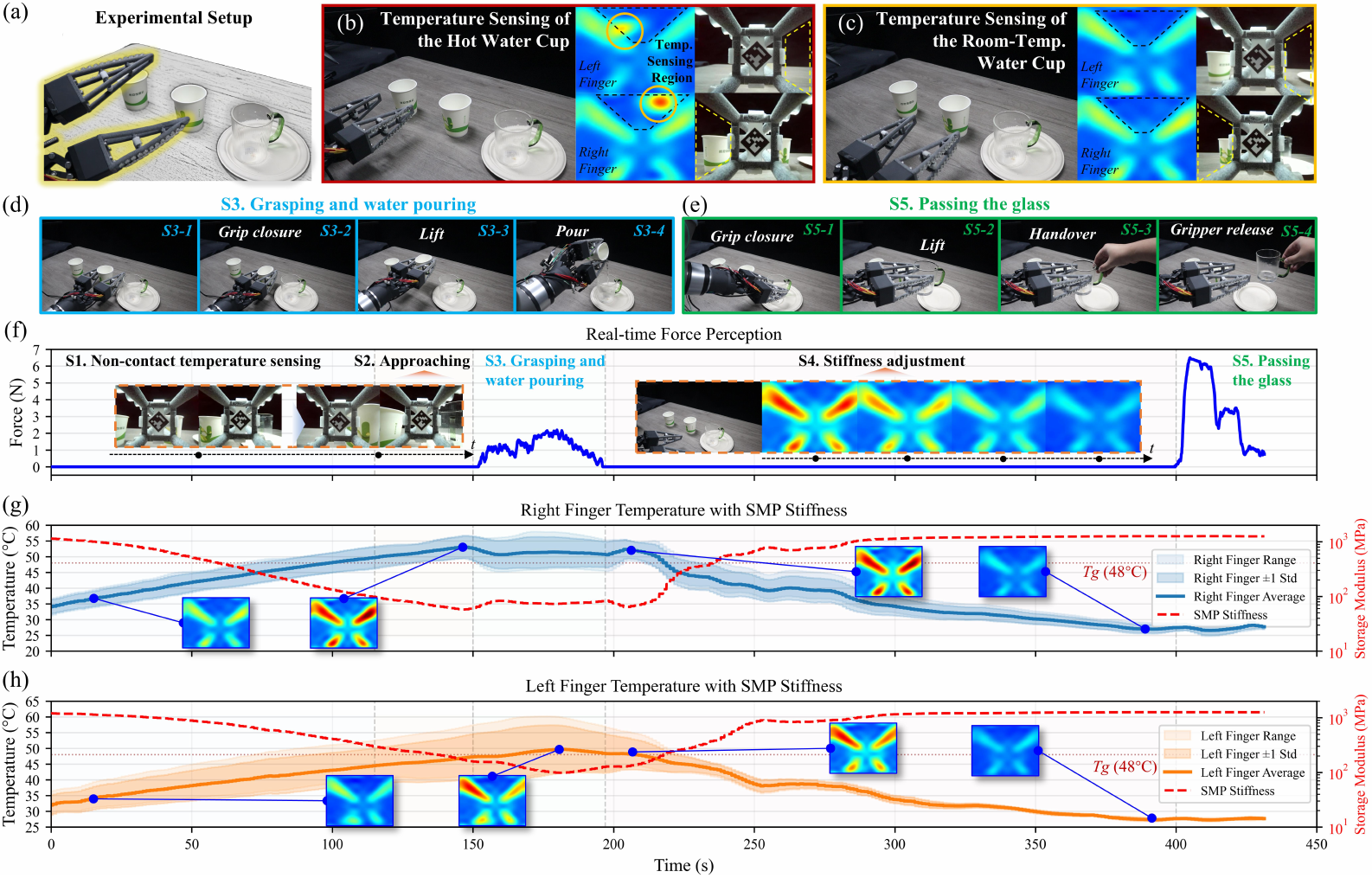}

\caption{
Integrated manipulation with variable stiffness and multimodal perception.
(a) Task setup.
(b--c) Non-contact thermal sensing. The infrared image identifies the temperature state of the selected cup.
(d) Compliant grasping and water-pouring sequence, including grip closure, lifting, and pouring.
(e) Stiff grasping sequence for handover, including grip closure, lifting, handover, and gripper release.
(f) Real-time force perception over the full manipulation sequence. The task is divided into five stages: S1, non-contact temperature sensing; S2, approaching; S3, compliant grasping and water pouring; S4, thermal stiffness adjustment; and S5, passing the glass.
(g--h) Temperature profiles of the right and left fingers and the corresponding temperature-derived SMP stiffness. The shaded regions indicate temperature variation across the finger, and the red dashed curves denote the estimated SMP storage modulus. The dotted line marks the glass-transition temperature $T_g \approx 48^\circ$C, around which the SMP stiffness changes rapidly.
}
    \label{fig:cup}
\end{figure*}

We present a system-level manipulation task to demonstrate the coordinated operation of variable stiffness, multimodal perception, and force control within a single continuous interaction sequence.
Unlike isolated evaluations, this experiment stresses the consistency of the sensing--actuation--control pipeline under changing object properties and interaction requirements.

As shown in Fig.~\ref{fig:cup}a, the scene contains two paper cups with water at different temperatures and an empty glass cup.
The task consists of three stages: non-contact identification of the room-temperature cup, compliant grasping and pouring, and stiff grasping with human handover.
These stages impose distinct and incompatible physical requirements, which cannot be satisfied by a fixed stiffness configuration.

During the perception stage (Fig.~\ref{fig:cup}b,c), infrared thermal sensing enables non-contact discrimination between the room-temperature and hot-water cups.
After target selection, the gripper approaches the paper cup using visual proximity cues from finger-mounted RGB cameras (Fig.~\ref{fig:cup}f S2).
Prior to grasping, finger temperature is increased to reduce stiffness, preparing the system for compliant interaction.

As illustrated in Fig.~\ref{fig:cup}d, the paper cup is grasped and lifted with a low target force of 1N, followed by water pouring.
Operating in a low-stiffness regime enlarges admissible deformation and reduces contact stress, improving robustness to misalignment and preventing damage to the deformable container.

The task then transitions to grasping the glass cup for handover (Fig.~\ref{fig:cup}e).
Before contact, finger stiffness is increased by lowering temperature, reconfiguring the gripper for higher load capacity and disturbance resistance.
The glass is then grasped and handed to a human, with force feedback maintaining stable interaction within the stiff regime at a 6N target force.

Figure~\ref{fig:cup}f shows the force profile over the entire sequence, clearly delineating perception, approach, compliant grasping, stiffness transition, and handover.
Figures~\ref{fig:cup}g,h report the corresponding temperature evolution and SMP stiffness states of both fingers.
Together, these results demonstrate that variable stiffness acts as a mechanism-level switch between interaction regimes,
while multimodal perception and force control ensure consistent and adaptive manipulation across heterogeneous task demands.

\section{Conclusion}

This paper targets stable long-duration grasp maintenance for soft grippers under thermo-mechanically coupled viscoelastic relaxation, where grasp forces decay during holding even with fixed geometry and actuation.
To address this challenge, we develop a variable-stiffness gripper with a relaxation-aware force perception and regulation framework enabled by multimodal sensing.
The system continuously senses and modulates interaction conditions across pre-contact approach, contact establishment, and prolonged holding, instead of relying solely on instantaneous contact feedback.

On the hardware side, a bio-inspired finger integrates passive compliance, thermally tunable stiffness, and switchable adhesion within a unified morphology, enabling multimode grasping from conformal contact to load-bearing holding while retaining a large deformation envelope for robust contact acquisition.
Base-mounted sensing preserves reliability under deformation: visible-light vision provides proximity cues and proprioceptive deformation tracking, while infrared thermography measures the temperature field associated with stiffness modulation.

On the modeling and control side, a temperature-coupled reduced viscoelastic force model separates instantaneous force from relaxation-induced decay.
A physics-informed learning scheme estimates the reduced quantities from multimodal observations, reconstructs relaxation-driven force evolution during holding, and provides an explicit compensation term for closed-loop regulation, enabling long-horizon force stabilization without embedded force sensors.

Experiments demonstrate robust contact acquisition, effective stiffness modulation, accurate force perception, and improved holding stability over fixed-aperture and instantaneous-only baselines.
Future work will extend the thermo-coupled relaxation compensation formulation to broader soft-hand morphologies, while further investigating the material mechanisms underlying thermally coupled relaxation.

\clearpage

\section*{Acknowledgments}
This work was supported in part by the National Natural Science Foundation of China Youth Program under Grant No. 52505041, the China Postdoctoral Science Foundation under Grant No. 2024M762814, the National Natural Science Foundation of China Youth Program under Grant No. 52305037, the Zhejiang Provincial Natural Science Foundation of China under Grant No. LD26E050001, and the ``Pioneer'' and ``Leading Goose'' R\&D Program of Zhejiang Province under Grant No. 2025C01072.

\bibliographystyle{unsrtnat}
\bibliography{references}

\end{document}